\documentclass[conference]{IEEEtran}

\usepackage{cite}
\usepackage{algorithmic}
\usepackage{textcomp}

\usepackage{amsmath,graphicx}
\usepackage{amssymb}
\usepackage{booktabs} 
\usepackage{multirow}
\usepackage[numbers]{natbib}
\usepackage{xcolor}
\usepackage{tcolorbox}

\usepackage{pifont}

\newcommand{\arrowrightdown}[2]{#1\textcolor{#2}{\boldmath$\downarrow$}}

\definecolor{c1}{cmyk}{0,0.6175,0.8848,0.1490} 
\definecolor{c2}{cmyk}{0.1127,0.6690,0,0.4431} 
\definecolor{c3}{cmyk}{0.3081,0,0.7209,0.3255} 
\definecolor{c4}{cmyk}{0.6765,0.2017,0,0.0667} 
\definecolor{c5}{cmyk}{0,0.8765,0.7099,0.3647} 
\definecolor{forestgreen}{HTML}{397727}

\definecolor{lightgreen}{RGB}{180,230,160} 
\definecolor{lightred}{RGB}{255,100,99} 

\usepackage{pgfplots}
\pgfplotsset{compat=1.17}
\usepackage{subcaption}
\usepackage{graphicx}
\begin{document}

\title{VITRO: Vocabulary Inversion for Time-series Representation Optimization\\
}

\author{
\IEEEauthorblockN{Filippos Bellos}
\IEEEauthorblockA{University of Michigan\\
Ann Arbor, MI, USA\\
Email: fbellos@umich.edu}
\and
\IEEEauthorblockN{Nam H. Nguyen}
\IEEEauthorblockA{Capital One\\
McLean, VA, USA\\
Email: nam.nguyen@capitalone.com}
\and
\IEEEauthorblockN{Jason J. Corso}
\IEEEauthorblockA{University of Michigan\\
Ann Arbor, MI, USA\\
Email: jjcorso@umich.edu}
}

\maketitle

\begin{abstract}
Although LLMs have demonstrated remarkable capabilities in processing and generating textual data, their pre-trained vocabularies are ill-suited for capturing the nuanced temporal dynamics and patterns inherent in time series. The discrete, symbolic nature of natural language tokens, which these vocabularies are designed to represent, does not align well with the continuous, numerical nature of time series data. To address this fundamental limitation, we propose \textit{VITRO}. Our method adapts textual inversion optimization from the vision-language domain in order to learn a new time series per-dataset vocabulary that bridges the gap between the discrete, semantic nature of natural language and the continuous, numerical nature of time series data. 
We show that learnable time series-specific pseudo-word embeddings represent time series data better than existing general language model vocabularies, with VITRO-enhanced methods achieving state-of-the-art performance in long-term forecasting across most datasets.
\end{abstract}
\section{Introduction}
\label{sec:intro}
Large Language Models (LLMs) have transformed
natural language processing (NLP), excelling in traditional NLP tasks like text generation but also showing promise in tasks that require complex and structured reasoning \cite{wei2022chain,bellos-etal-2024-large}.
Their impact has extended beyond NLP, contributing to rapid advancements in computer vision and other signal processing applications through the development of multimodal models that can process and integrate information from various modalities such as text, images, and audio.
This versatility has naturally led the community to explore their potential in time series forecasting, a fundamental capability in numerous real-world dynamic systems~\cite{jin2023survey} including energy load management~\cite{liu2023sadi}, climate modelling~\cite{schneider1974climate}, traffic forecasting~\cite{DBLP:journals/corr/abs-2202-03630},   \textit{etc}.  Traditionally, these forecasting tasks have required extensive domain expertise and task-specific model designs, an approach that stands in contrast to LLMs, which demonstrate strong performance across diverse tasks with minimal examples, often in few-shot or zero-shot scenarios ~\cite{brown2020language,kojima2205large}. 
\begin{figure}[htbp] 
  \centering
  \hspace*{-0.35cm} 
  \includegraphics[width=0.5\textwidth]{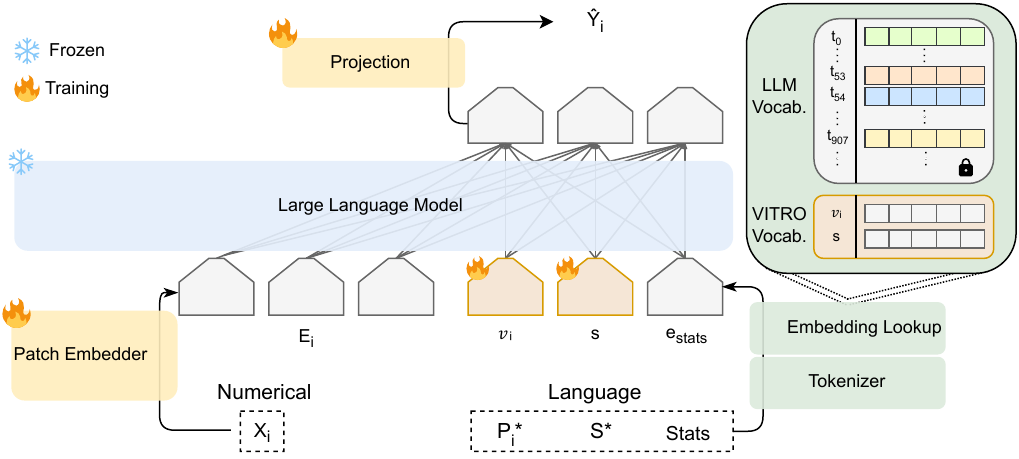}\vspace{-2mm} %
  \caption{
  VITRO optimizes learnable pseudo-word embeddings $v_i$ for each time series instance $X_i$ and a shared dataset embedding $s$ to construct a new data-centric time series vocabulary tailored for forecasting. Time series are normalized, patched, and embedded. These patch embeddings $E_i$ serve as prompts to guide the optimization of pseudo-words. The composite representation, including statistical features $e_{stats}$, is fed into a frozen LLM, whose output is projected to generate forecasts $\hat{Y}_i$.
  }
  \label{pipeline} 
  \vspace{-6mm}
\end{figure}
This contrast underscores the need to consider if and how the pre-trained knowledge and generalization capabilities of LLMs can be fully harnessed to perform accurate time series forecasting without fine-tuning the underlying model.

Time-LLM~\cite{jin2024timellm} and TEST~\cite{sun2024test} attempt to address this challenge by reprogramming the input time series into text prototype representations and using textual prompts to provide additional context. Crucially, these methods enable the LLM to perform time series forecasting while keeping the pre-trained model completely frozen, thus fully leveraging the model's pre-trained capabilities. Other methods, such as OFA \cite{zhou2023one} and $S^2$IP-LLM \cite{pan2024sipllm}, also investigate the use of pre-trained LLMs for time-series forecasting. However, they require partial fine-tuning of the underlying language model to achieve good performance, potentially limiting their ability to fully exploit the LLM's pre-trained knowledge. 

Despite the promise in these methods, they are still limited by the existing LLM vocabulary they rely on, which fails to capture the nuanced patterns and characteristics specific to time series data. This limitation naturally raises the question: 
\textit{Is there a better way to represent time series data than using the general-purpose vocabulary of LLMs, in order to leverage the inherent capabilities of LLMs for effective time series forecasting?}

To address this question, we propose VITRO, a new method that, as depicted in Fig.\ref{pipeline}, constructs a time series specific vocabulary by learning unique pseudo-words for each time series instance in a dataset, inspired by the concept of textual inversion~\cite{gal2022textual}. In addition, VITRO optimizes a shared embedding, able to capture the domain specific dataset information, which in many real-world applications may not always be available or informative. 

Intuitively, this method bridges the gap between LLMs and time series data by creating a 
vocabulary that encodes time series information in a format interpretable by the language model, while at the same time capturing the subtle variations in time series.

We demonstrate that VITRO can be leveraged across different forecasting approaches and LLM architectures, showcasing its potential for broad application in the field of time series forecasting.
Quantitative experiments show state-of-the-art performance for the methods that leverage VITRO, while qualitative analysis reveals that our learned vocabulary exhibits distinct patterns in attention weights and embedding distributions, indicating successful specialization for time series tasks.
\section{Method}
\subsection{Problem Formulation and Overview}

We formulate our problem of Vocabulary Inversion for Time Series Representation Optimization as follows. Let $X \in \mathbb{R}^{N \times T}$ denote the time series data consisting of $N$ different 1-dimensional variables across a lookback window of $T$ time steps, where the $i$-th series is denoted as $X_i \in \mathbb{R}^{1 \times T}$. We aim to learn a  new time series data-centric vocabulary that will allow a large language model $f(\cdot)$ to better understand the input time series in order to more accurately predict the next $\tau$ time steps based on the input window T.
Let $Y \in \mathbb{R}^{N \times \tau}$ denote the ground truth values for the next $\tau$ time steps, and $\hat{Y} \in \mathbb{R}^{N \times \tau}$ represent the corresponding predictions. The overall objective is to minimize the mean square errors between $Y$ and $\hat{Y}$, defined as:
\begin{equation}
\frac{1}{\tau} \sum_{t=1}^{\tau} |Y_t - \hat{Y}_t|^2_F.
\end{equation}

To solve this problem, we are inspired by the approach of textual inversion \cite{gal2022textual} from the text-to-image diffusion model literature—a simple yet powerful technique in low-shot image
generation that learns a common concept in given images as a single token in text embedding space. Motivated by its success, we aim to learn different concepts-representations of time series data as text embeddings and use them to develop a new time series forecasting vocabulary, which we hypothesize will represent time series data better than the existing general natural language model vocabulary.

Our method consists of two stages. The first stage optimizes a specialized vocabulary tailored for time series forecasting. This stage captures patterns across an entire dataset, creating a rich vocabulary that reflects the temporal dynamics and variations inherent in the time series. The primary goal here is to establish a comprehensive representation rather than immediate forecasting accuracy. In the second stage, we use this specialized vocabulary for the actual forecasting tasks. This stage benefits from the broad context learned in the first stage, applying it to individual time series instances to enhance forecasting performance.  The two-stage approach ensures that our vocabulary is informed by the full dataset. It allows us to first establish a strong foundational representation before focusing on specific forecasting tasks.

\subsection{Stage 1: Vocabulary Inversion for Time Series}

LLMs begin with a text processing step where each word or sub-word in an input string is converted to a token, which is an index in some pre-defined dictionary. Each token is then linked to a unique embedding vector that can be retrieved through an index-based embedding lookup.

We choose this embedding space as the target for vocabulary inversion. Specifically, let a time series dataset $D = \{X_1, X_2, \ldots, X_n\}$, where $n$ represents the number of time series instances in a dataset and each time series instance $X_i \in \mathbb{R}^{1 \times T}$ represents a time series segment of length $T$ (lookback window). 
For this dataset, we designate a set of placeholder strings, $P^* = \{P^*_1, P^*_2, \ldots, P^*_n\}$, where each $P^*_i$ represents a unique time series instance $X_i$. Additionally, we introduce a placeholder $S^*$ shared per dataset which represents the entire dataset-domain information.

Concretely, our approach involves an iterative optimization process for:
\begin{itemize}
\item A corresponding embedding $v_i$ for each $P^*_i$, effectively expanding the LLM's vocabulary with $n$ new ``words'' that encode time series information.
\item A corresponding embedding $s$ for $S^*$, that encodes the domain information for $D$ as a whole.
\end{itemize}
We start by randomly initializing each pseudo-word embedding $v_i$ and associating the placeholder $P^*_i$ to it. Similarly, we initialize the shared embedding $s$ and associate it with $S^*$. The optimization process then runs for a fixed number of iterations, optimizing these embeddings to better represent the time series forecasting data.
To condition the generation process, we utilize a small set of text prompt templates containing these placeholders, such as ``The time series is $P_i^*$'' or ``Forecast the next steps of $P_i^*$''.

As shown in Fig. \ref{pipeline}, we essentially intervene in the LLM's embedding process. This allows us to ``inject" a rich set of time series concepts into the LLM's vocabulary, each 
reflecting 
a specific instance in our dataset. The same happens with the shared embedding, which "injects" general domain information.

\subsubsection{Model Pipeline}

Following convention, for each input time series \(X_i \in \mathbb{R}^T\), we first apply reversible instance normalization (RevIN)~\cite{kim2022reversible} to mitigate distribution shift: \(\tilde{X}_i = \text{RevIN}(X_i)\). We then divide \(\tilde{X}_i\) into \(P\) overlapping or non-overlapping patches of length \(L_p\): \(X_{P,i} \in \mathbb{R}^{P \times L_p}\)~\cite{nie2022time}, where \(P = \left\lfloor\frac{T-L_p}{S}\right\rfloor + 2\), and \(S\) is the horizontal sliding stride.

To obtain the final embeddings, we apply a linear transformation to each patch: \(E_i = W_e X_{P,i} + b_e\), where \(E_i \in \mathbb{R}^{P \times d}\), \(W_e \in \mathbb{R}^{d \times L_p}\) is a learnable weight matrix, \(b_e \in \mathbb{R}^d\) is a learnable bias vector, and \(d\) is the embedding dimension of the target LLM.

\textbf{Patch Embeddings as Prompts.} We leverage the patch embeddings $E_i$, as a composite prompt to guide the LLM's processing of time series data and the optimization of pseudo-words $v_i$, and shared embedding $s$. This approach is inspired by recent advancements showing that non-textual data modalities can be effectively integrated as prefixes in prompts to facilitate reasoning~\cite{tsimpoukelli2021multimodal}. 
In our case, the patch embeddings $E_i$ serve as a numerical representation of the time series, while the pseudo-words $v_i$ and shared embedding $s$ provide learnable, text-like anchors for the model.

For each pseudo-word $P_i^*$, we learn a corresponding embedding vector $v_i \in \mathbb{R}^d$. We pass $P_i^*$ through the LLM's tokenizer to obtain a token representation which we associate in the LLM's embedding lookup table with our learnable embedding $v_i$. Accordingly, we associate the shared word $S^*$ with the learned shared embedding $s$ in the embedding lookup process.

We then concatenate \({E}_i\), $v_i$ and $s$ along with certain statistics we calculate for $X_i$, and we feed them through the frozen LLM to obtain the last hidden layer output \(h_i \in \mathbb{R}^h\): \(h_i = f([E_i; u_i; s; e_{stats}])\), where $f(\cdot)$ denotes the frozen LLM and $;$ denotes concatenation.

The last hidden layer output \(h_i\) is passed through a learnable linear layer \(g(\cdot)\) to generate the forecasted values \(\hat{Y}_i\): \(\hat{Y}_i = g(h_i) = W \times h_i + b\), where \(W \in \mathbb{R}^{\tau \times h}\) and \(b \in \mathbb{R}^\tau\) are the learnable weights and bias.

\subsubsection{Optimization Objective}

Our optimization objective is to minimize the loss $\mathcal{L}$ between the forecasted values $\hat{Y}_i$ and the ground truth future values $Y_i$ for each time series instance $X_i$:

\begin{align}
\mathcal{L}_{MSE} &= \frac{1}{\tau}\left\| {Y}_i - \left( g \left( f\left( {E}_i ; \text{``The time series} \right. \right. \right. \right. \nonumber \\
& \left. \left. \left. \text{is } [P_i^*], \text{The dataset is } [S^*] \text{''} ; \text{Statistics}({X}_i) \right) \right) \right\|_F^2  
\end{align}

We optimize the shared embedding $s$, pseudo-word embeddings $V = {v_1, v_2, ..., v_n}$
and all other learnable parameters $\theta$ (including $W_e$, $b_e$, $W$ and $b$) to minimize the total loss:

\begin{equation}
s^*, V^*, \theta^* = \arg \min_{s, V, \theta} \sum_{i=1}^n \mathcal{L}_{MSE}
\end{equation}

\subsection{Stage 2: Time Series Forecasting with Learned Vocabulary}
Stage 2 of our method focuses on leveraging the learned vocabulary V for time series forecasting. We present two approaches using different LLM architectures (i.e. GPT2~\cite{radford2019language} and LLaMa~\cite{touvron2023llama}) demostrating that the effectiveness of VITRO is not restricted to one type of LLM or LLM-based method: a similarity-based selection method that directly utilizes the vocabulary, and TimeLLM's \cite{jin2024timellm} attention-based approach that allowing us to assess VITRO's benefits compared to the standard LLM vocabulary within the TimeLLM method, which is the current state of the art method that utilizes a frozen pretrained LLM. 

\textbf{Similarity-based Selection (Sim)} For computational efficiency, instead of using the word embeddings from the full vocabulary $V$, we first derive a reduced set of core lexicon embeddings $C$ using a linear mapping function $h(\cdot)$: $C = h(V) = W_v V + b_v$, where $C \in \mathbb{R}^{n' \times d}$, $n'$ is the number of core lexicon embeddings with $n' \ll n$, $W_v \in \mathbb{R}^{n' \times n}$ and $b_v \in \mathbb{R}^{n'}$ are learnable parameters. 
For each patch embedding $E_i$ and each core lexicon embedding $c_m \in C$, we compute the cosine similarity: $\text{sim}(E_i, c_m) = (E_i \cdot c_m) / (\|E_i\| \|c_m\|)$. From these $n'$ core lexicon embeddings, we then select the top $k$ embeddings with the highest similarity scores, where $k < n'$. This similarity-based ranking and selection ensures that we identify the most relevant core lexicon embeddings for each patch, while maintaining computational efficiency by operating in a reduced similarity space ($n'$ instead of $n$).

We then form an augmented embedding for each patch: $\hat{e}_i = [E_i; 
c_1; c_2; ...; c_k; s; e_{stats}]$, that will serve as input to the frozen pre-trained LLM, which in this case is GPT2.

\textbf{Attention-based approach}
As mentioned, we also employ TimeLLM's~\cite{jin2024timellm} attention-based approach , demonstrating the improved results (in Section \ref{results}) when using our optimized vocabulary over the existing one. This method involves a multi-head cross-attention mechanism between patch embeddings and our optimized vocabulary, allowing the model to dynamically select relevant information. 

Concretely, we employ a multi-head cross-attention layer.
For each head $h = \{1, ..., H\}$, we define the Query matrices as $ Q_h^{(i)} = E_iW_h^Q $, the Key matrices as $K_h^{(i)} = CW_h^K $ and the Value matrices as $V_h^{(i)} = CW_h^V$,
where $W_h^Q, W_h^K, W_h^V \in \mathbb{R}^{d \times d_h}$, and $d_h = d/H$.

The attention operation for each head is: \\
$Z_h^{(i)} = Softmax\left(\frac{Q_h^{(i)}K_h^{(i)\top}}{\sqrt{d_h}}\right)V_h^{(i)}$. Aggregating $Z_h^{(i)} \in \mathbb{R}^{P \times d_h}$ across all heads yields $Z^{(i)} \in \mathbb{R}^{P \times d}$, which is then passed through the frozen LLM, which in this case is Llama-7B, along with the optimized shared embedding that encapsulates the domain information, and the calculated statistics.

\begin{table*}[!h]

\footnotesize
\centering
\begin{tabular}{ll|ll|ll|ll|ll|ll|ll|ll}
\hline\hline
\multicolumn{2}{c|}{Methods} &
  \multicolumn{2}{c|}{\textbf{VITRO-Sim}} &
  \multicolumn{2}{c|}{\textbf{Sim}}&
  \multicolumn{2}{c|}{\textbf{VITRO-TimeLLM}} &
  \multicolumn{2}{c|}{\textbf{TimeLLM}} &
  \multicolumn{2}{c|}{\textbf{$S^2$IP-LLM}} &
  \multicolumn{2}{c|}{\textbf{PatchTST}} &
  \multicolumn{2}{c}{\textbf{Dlinear}}    \\ \hline
\multicolumn{2}{c|}{Metric} &
  \multicolumn{1}{c}{MSE} &
  \multicolumn{1}{c|}{MAE} &
  \multicolumn{1}{c}{MSE} &
  \multicolumn{1}{c|}{MAE} &
  \multicolumn{1}{c}{MSE} &
  \multicolumn{1}{c|}{MAE} &
  \multicolumn{1}{c}{MSE} &
  \multicolumn{1}{c|}{MAE} &
  \multicolumn{1}{c}{MSE} &
  \multicolumn{1}{c|}{MAE} &
  \multicolumn{1}{c}{MSE} &
  \multicolumn{1}{c|}{MAE} &
  \multicolumn{1}{c}{MSE} &
  \multicolumn{1}{c}{MAE} \\ \hline

\multicolumn{2}{l|}{\textbf{ETTh1}} & \arrowrightdown{\textbf{0.412}}{lightgreen} & \arrowrightdown{\textbf{0.430}}{lightgreen} & 0.442 & 0.449 & \arrowrightdown{\underline{0.416}}{lightgreen} & \arrowrightdown{\underline{0.437}}{lightgreen} & 0.437 & 0.450  & 0.425 & 0.440 & 0.444 & 0.453 & 0.418  & 0.439  \\
\hline

\multicolumn{2}{l|}{\textbf{ETTh2}}& \arrowrightdown{\underline{0.351}}{lightgreen} & \arrowrightdown{\textbf{0.393}}{lightgreen} & 0.370 & 0.402 & \arrowrightdown{\textbf{0.349}}{lightgreen} & \arrowrightdown{\underline{0.395}}{lightgreen} & 0.360 & 0.400  & 0.358 & 0.403 & 0.381 & 0.411 & 0.502 & 0.481 \\
\hline

\multicolumn{2}{l|}{\textbf{ETTm1}}& \arrowrightdown{0.353}{lightgreen} & \arrowrightdown{\textbf{0.380}}{lightgreen} &  0.365 &  0.388 & \arrowrightdown{\underline{0.352}}{lightgreen} & \arrowrightdown{0.387}{lightgreen} & 0.367 & 0.396 & \textbf{0.347} & \underline{0.382} & 0.363 & 0.391 & 0.357 & 0.389  \\
\hline

\multicolumn{2}{l|}{\textbf{ETTm2}} & \arrowrightdown{\textbf{0.260}}{lightgreen} & \arrowrightdown{\underline{0.323}}{lightgreen} & 0.284 & 0.332 & \arrowrightdown{0.263}{lightgreen} & \arrowrightdown{\textbf{0.321}}{lightgreen} & 0.264 & 0.325 & \underline{0.261} & 0.326 & 0.267 & 0.325 & 0.275 & 0.340 \\
\hline

\multicolumn{2}{l|}{\textbf{Weather}} & \arrowrightdown{0.230}{lightgreen} & \arrowrightdown{0.268}{lightgreen} & 0.233 & 0.273 & \arrowrightdown{\textbf{0.225}}{lightgreen} & \arrowrightdown{\textbf{0.263}}{lightgreen} & \underline{0.227} & 0.265 & 0.229 & 0.267 & \textbf{0.225} & \underline{0.264} & 0.248 & 0.300 \\
\hline

\multicolumn{2}{l|}{\textbf{Electricity}} & \arrowrightdown{\textbf{0.161}}{lightgreen} & \arrowrightdown{\underline{0.258}}{lightgreen} & \underline{0.165} & 0.261 & \arrowrightdown{0.166}{lightgreen} & \arrowrightdown{0.267}{lightgreen} & 0.168  &  0.270 &    0.167 & 0.263 & \textbf{0.161} & \textbf{0.252} & 0.166 & 0.263  \\
\hline

\multicolumn{2}{l|}{\textbf{Traffic}} & \arrowrightdown{\underline{0.399}}{lightgreen} & \arrowrightdown{\underline{0.276}}{lightgreen} &  0.402 & 0.279 & \arrowrightdown{0.408}{lightgreen} & \arrowrightdown{0.306}{lightgreen} &  0.410 &  0.310 & 0.418 & 0.303 & \textbf{0.390} & \textbf{0.263} & 0.433 & 0.295 \\

\bottomrule

\end{tabular}
\caption{Long-term forecasting results for \{96, 192, 336, 720\} horizons. A lower value indicates a better performance. 
 All results are averaged from four forecasting horizons\{96, 192, 336, 720\}. \arrowrightdown{Arrows}{lightgreen} indicate positive impact of VITRO compared to existing LLM vocabulary. \textbf{Bold}: best results. \underline{Underlined}: second best. We reproduced TimeLLM, $S^2$IP-LLM, PatchTST, Dlinear results using their official open-source implementations.}
\label{tab:few10}
\end{table*}
\section{Experiments}
\label{results}

In our experimental evaluation, we compare the effectiveness of VITRO vocabularies against the general natural language LLM vocabularies within our Stage 2 framework, demonstrating the advantages of our optimized time series representation. We benchmark VITRO-enhanced methods against other LLM-based methods and traditional time series forecasting baselines on the task of long-term forecasting. We also provide a qualitative analysis of VITRO and existing LLM vocabularies. For all baselines, we adhere to the experimental configurations outlined by~\cite{wu2022timesnet}, utilizing their unified pipeline\footnote{https://github.com/thuml/Time-Series-Library}.

\textbf{Baselines.} 
The baselines include the best performing approaches based on LLMs, \textit{i.e.}, Time-LLM~\cite{jin2024timellm} whose method is used as variant of our stage 2, 
and $S^2$IP-LLM~\cite{pan2024sipllm} even though this method partly finetunes the backbone model. We also include the best performing Transformer-based  and non-Transformer methods, \textit{i.e.} PatchTST~\cite{nie2022time} and DLinear~\citep{Zeng2022AreTE}.

\subsection{Long-term Forecasting}

\textbf{Setup.} We evaluate the effectiveness of VITRO across 7 public datasets: Weather, Electricity, Traffic, and four ETT datasets (\textit{i.e.}, ETTh1, ETTh2, ETTm1, and ETTm2), which have been widely adopted as benchmarking datasets for long-term forecasting models. The input time series length is 512 and we evaluate the performance on four different horizons $\{ 96, 192, 336, 720\}$. The evaluation metrics include the mean square error (MSE) and the mean absolute error (MAE).

\textbf{Results.} Our results are shown in TABLE \ref{tab:few10}. When we replace the existing general-purpose vocabulary with VITRO's learned vocabulary in both the Time-LLM approach and our similarity-based method, we observe consistent improvements across all 7 datasets tested for both MSE and MAE metrics. The impact is particularly pronounced for the ETTh1, ETTh2 and  ETTm1 datasets. 
When compared to state-of-the-art methods, our VITRO-enhanced approaches consistently outperform across most datasets. Specifically, for the MAE metric our methods outperform the LLM-based method ($S^2$IP-LLM) in all 7 datasets while for MSE in 6 out of 7 datasets. Comparing against the transformer based method (PatchTST) VITRO performs better in 5 out of 7 datasets for the MAE metric, 4 out of 7 for MSE while achieving the same result for the same metric in 2 datasets (i.e. Electricity, Weather). Finally, we outperform the non-transformer method (Dlinear) in all datasets for both metrics.
\section{Qualitative Analysis}

\begin{figure}[t] 
  \centering
  \includegraphics[width=0.45\textwidth]{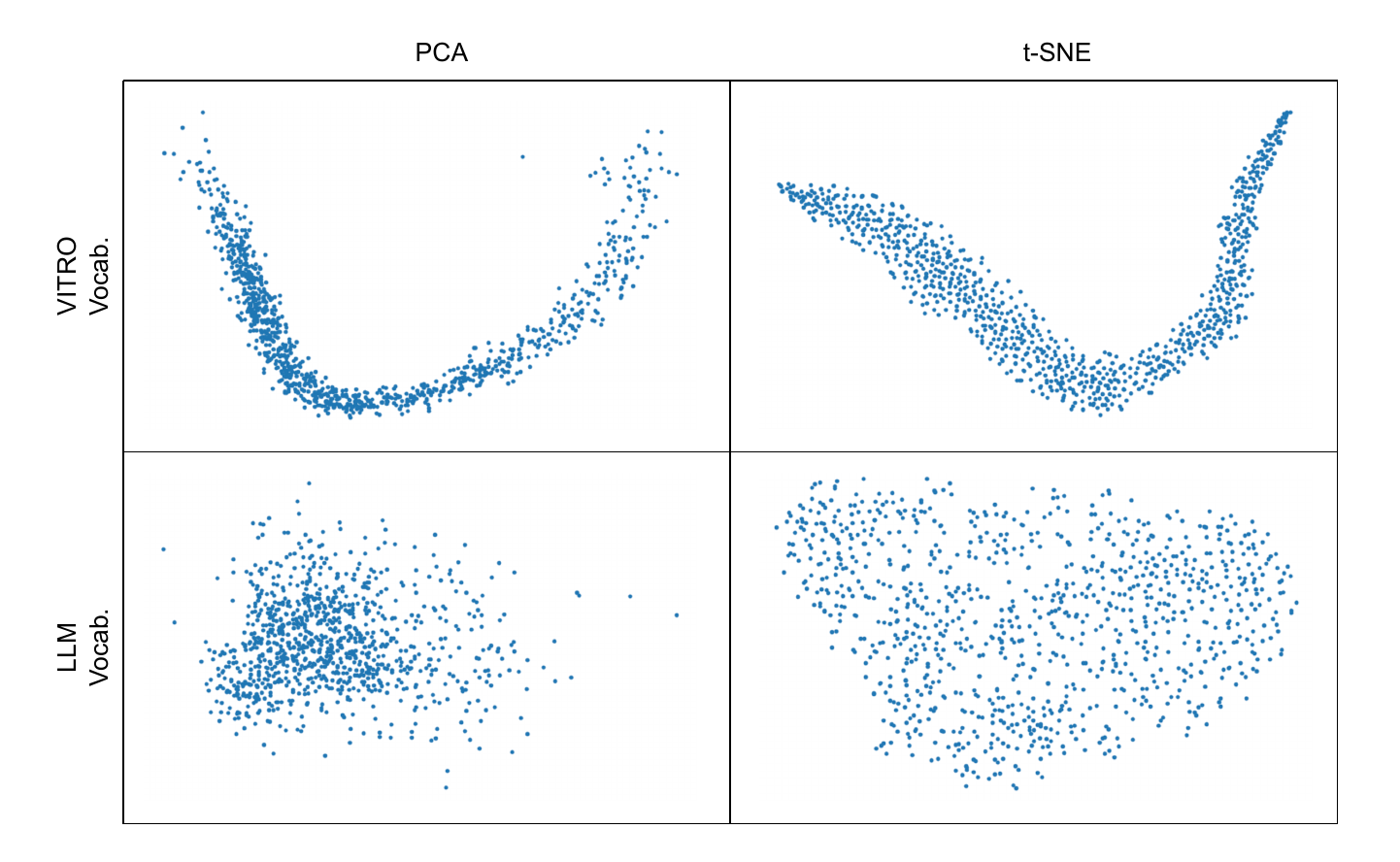}\vspace{-3mm} %
  \caption{PCA and t-SNE visualizations of VITRO and existing general-purpose vocabulary embedding space.}
  \label{viz} 
  \vspace{-5mm}
\end{figure}

\begin{figure}[t] 
  \centering
  \includegraphics[width=0.45\textwidth]{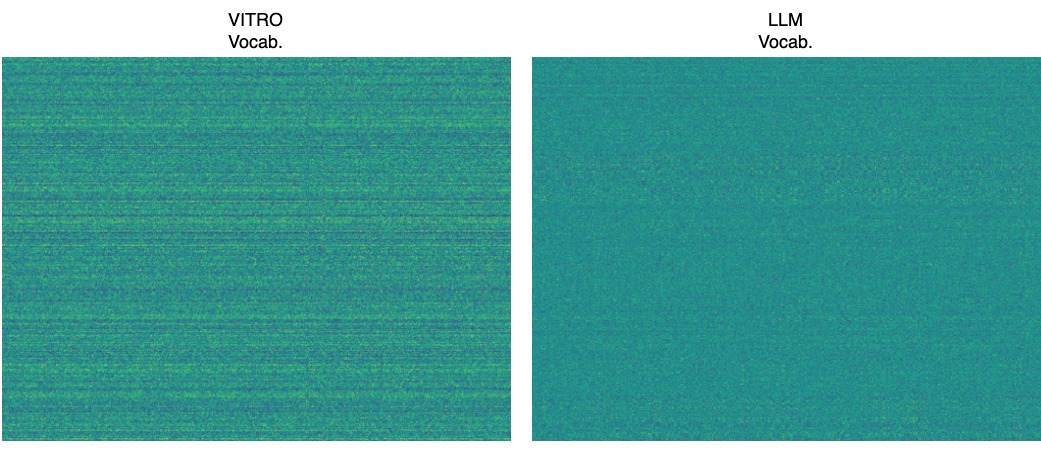}\vspace{-3mm} %
  \caption{VITRO and LLM existing vocabularies heatmaps. Each row corresponds to a word in the vocabulary, the y-axis represents the index of the word, and the x-axis denotes the embedding dimensions. Brighter colors indicate higher values.}
  \label{attention} 
  \vspace{-6mm}
\end{figure}

Figures \ref{viz} and \ref{attention} reveal the impact of the specialized nature of the new vocabulary for time series tasks, contrasting with the general-purpose characteristics of existing vocabularies. 

 In Fig. \ref{attention}, the heatmaps, generated by the attention-based approach of stage 2 (TimeLLM approach), for the VITRO vocabulary show distinct horizontal striping patterns. This suggests that certain vocabulary elements are consistently more important across different parts of the input sequence, indicating that our vocabulary has captured some general features and underlying structures in time series data applicable across various time steps. In contrast, the existing vocabulary's weights show a more uniform distribution, suited for more general language tasks. Figure \ref{viz}'s PCA and TSNE visualizations further support this distinction: the new vocabulary forms a U-shaped manifold,  suggesting a robust and structured embedding space, which indicates a specialized representation of time series concepts, while the existing vocabulary reveals a diffuse, circular distribution 
 typical of general-purpose language embeddings.

\section{Conclusion and Future Work}
VITRO demonstrates significant potential in enhancing LLMs for time series forecasting by learning a time series data-centric vocabulary through vocabulary inversion. Our results consistently show that time series forecasting accuracy can be improved by replacing the LLM's general-purpose vocabulary with our VITRO-optimized one. However, as an iterative optimization-based method, VITRO's computational cost may limit its application in larger datasets.
Future research directions will explore further optimization of the vocabulary learning process, extending VITRO to other time series tasks beyond forecasting, and integrating VITRO with other LLM-based methods (e.g. $S^2$IP-LLM). 

\bibliographystyle{IEEEbib}
\bibliography{strings,refs}

\end{document}